\documentclass[runningheads]{llncs}
\usepackage{graphicx}
\usepackage{tikz}
\usepackage{comment}
\usepackage{amsmath,amssymb} % define this before the line numbering.
\usepackage{color}

\usepackage{epsfig}
\usepackage{graphicx}
\usepackage{amsmath}
\usepackage{amssymb}
\usepackage{graphicx}
\usepackage{makecell}
\usepackage{epstopdf}
\usepackage{hyperref}	

\hypersetup{colorlinks,%				% Reference color setup	
	linkcolor=red,%
	citecolor=blue}
\usepackage{amsfonts}
\usepackage{color}
\usepackage{multirow}
\usepackage{setspace}
\usepackage{booktabs}
\usepackage{enumerate}
\usepackage[linesnumbered,ruled,vlined]{algorithm2e}
\usepackage{amssymb}
\usepackage{mathrsfs}
\usepackage{setspace}
\usepackage{colortbl} 
\usepackage{arydshln} 
\usepackage{amssymb}
\usepackage{array}
\usepackage{mathrsfs}
\usepackage{pifont}  % ding
\usepackage{orcidlink}

\usepackage[accsupp]{axessibility}  % Improves PDF readability for those with disabilities.
\usepackage{lipsum}

\definecolor{mygray}{gray}{.95}

\begin{document}
% \linenumbers
\pagestyle{headings}
\mainmatter

\title{Panoramic Human Activity Recognition} % Replace with your title

\begin{comment}
\titlerunning{ECCV-22 submission ID \ECCVSubNumber} 
\authorrunning{ECCV-22 submission ID \ECCVSubNumber} 
\author{Anonymous ECCV submission}
\institute{Paper ID \ECCVSubNumber}
\end{comment}
%******************

%% CAMERA READY SUBMISSION
%%\begin{comment}
%\titlerunning{Panoramic Human Activity Recognition}
%% If the paper title is too long for the running head, you can set
%% an abbreviated paper title here
%%
%\author{Ruize Han$^*$\inst{1} \and
%Haomin Yan$^*$\inst{1} \and
%Jiacheng Li$^*$\inst{1} \and \\
%Songmiao Wang \inst{1} 
%\and Wei Feng \inst{1}
%\and Song Wang \inst{2}}
%%
%\authorrunning{R. Han et al.}
%% First names are abbreviated in the running head.
%% If there are more than two authors, 'et al.' is used.
%%
%\institute{Tianjin University, China \and
%University of South Carolina, Columbia, USA \\
%\email{han\_ruize@tju.edu.cn \\{$^*$ Equal Contribution}} }
%%\url{http://www.springer.com/gp/computer-science/lncs} \and
%%ABC Institute, Rupert-Karls-University Heidelberg, Heidelberg, Germany\\
%%\email{\{abc,lncs\}@uni-heidelberg.de}}
%%\end{comment}
%%******************

%\begin{comment}
\titlerunning{Panoramic Human Activity Recognition}
\author{Ruize Han\inst{1}\orcidlink{0000-0002-6587-8936} \and
	Haomin Yan\inst{1}$^*$\and
	Jiacheng Li\inst{1}$^*$\orcidlink{0000-0002-2078-5998} \and \\
	Songmiao Wang\inst{1}
	\and Wei Feng\inst{1}$^\dagger$\orcidlink{0000-0003-3809-1086} {}
	\and Song Wang\inst{2}$^\dagger$\orcidlink{0000-0003-4152-5295}}
\authorrunning{R. Han et al.}
\institute{Intelligence and Computing College, Tianjin University, China 
	\and University of South Carolina, Columbia, USA 
	\\
	\email{\{han\_ruize, yan\_hm, threeswords, smwang, wfeng\}@tju.edu.cn, songwang@cec.sc.edu }}
%\end{comment}
%******************

%% CAMERA READY SUBMISSION
%%\begin{comment}
%% If the paper title is too long for the running head, you can set
%% an abbreviated paper title here
%%
%\author{First E. van Author\inst{1}\orcidlink{0000-1111-2222-3333}\index{van Author, First E.} \and
%	Second Author\inst{2,3}\orcidlink{1111-2222-3333-4444} \and
%	Third Author\inst{3}\orcidlink{2222--3333-4444-5555}}
%% First names are abbreviated in the running head.
%% If there are more than two authors, 'et al.' is used.
%%
%\institute{Princeton University, Princeton NJ 08544, USA \and
%	Springer Heidelberg, Tiergartenstr. 17, 69121 Heidelberg, Germany
%	\email{lncs@springer.com}\\
%	\url{http://www.springer.com/gp/computer-science/lncs} \and
%	ABC Institute, Rupert-Karls-University Heidelberg, Heidelberg, Germany\\
%	\email{\{abc,lncs\}@uni-heidelberg.de}}
%%\end{comment}
%%******************

\maketitle

\newcommand\blfootnote[1]{%
	\begingroup
	\renewcommand\thefootnote{}\footnote{#1}%
	\addtocounter{footnote}{-1}%
	\endgroup
}

\begin{abstract}
To obtain a more comprehensive activity understanding for a crowded scene, in this paper, we propose a new problem of panoramic human activity recognition (PAR), which aims to simultaneously achieve the the recognition of individual actions, social group activities, and global activities.
This is a challenging yet practical problem in real-world applications.
To track this problem, we develop a novel hierarchical graph neural network to progressively represent and model the multi-granular human activities and mutual social relations for a crowd of people.
We further build a benchmark to evaluate the proposed method and other related methods. 
Experimental results verify the rationality of the proposed PAR problem, the effectiveness of our method and the usefulness of the benchmark. We have released the source code and benchmark  to the public for promoting the study on this problem. 
\end{abstract}

%%%%%%%%% BODY TEXT
\section{Introduction}

\begin{figure*}[ht!] 
	\centering 
	\includegraphics[width=0.975\linewidth]{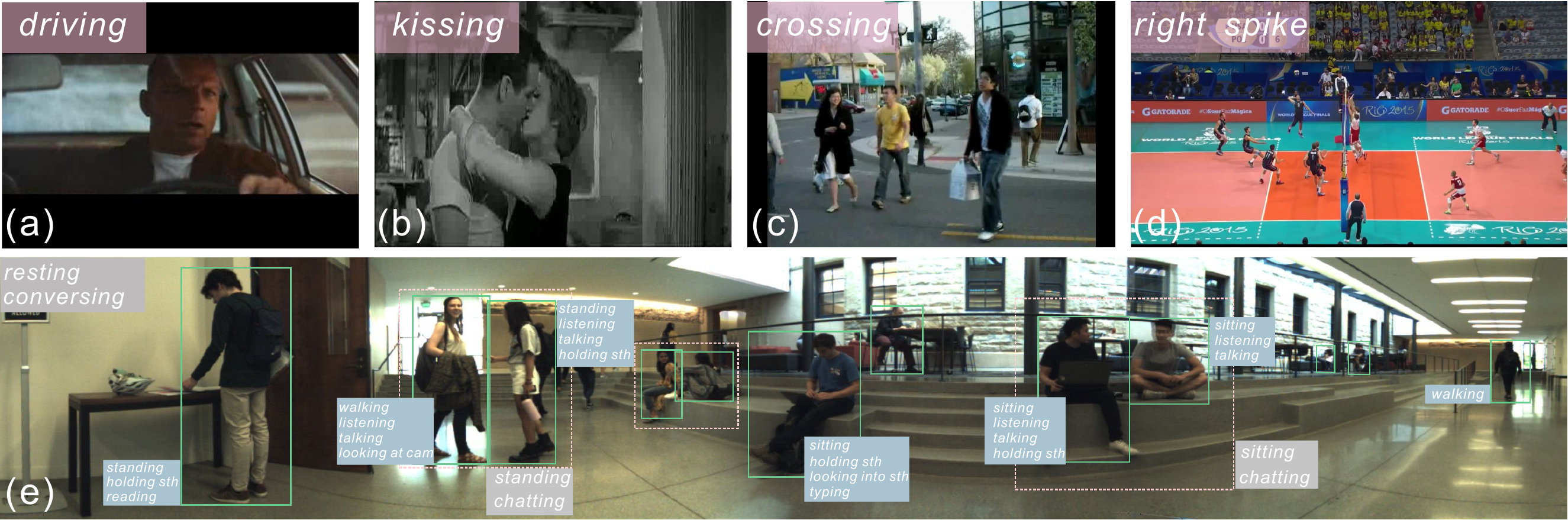}
	\caption{Examples of different types of activity recognition -- (a) Action recognition. (b) Interaction recognition. (c-d) Group activity recognition. (e) The proposed panoramic human activity recognition (PAR), in which the labels beside each human bounding box (green) denote the individual actions, beside each human group box (pink dotted) denote the social activities, and the label at the top left corner of the image denotes the global activity.}
	\label{fig:example}  
\end{figure*}

\blfootnote{$^*$Equal Contribution. $^\dagger$Corresponding Authors.}Video-based human activity understanding is an important computer vision task, which has various practical applications in real world, e.g., video surveillance and social scene analysis~\cite{zhao2020human,han2020cip}.
%Therefore, a large number of works have been proposed for this task. 
In the past decade, this challenging task has been drawing much research interest in the computer-vision community. 
As shown in Fig.~\ref{fig:example}, previous works on human activity recognition can be divided into three categories.  1) Human action recognition aims to recognize the action categories of individual persons in a video~\cite{Sf-net,zhang2020multi,slowfast}. 2) Human interaction recognition is proposed to recognize the human-human interactions~\cite{2019Analyzing}. 3) Group activity recognition is to recognize the overall activity of a group of people~\cite{Ibrahim_2018_ECCV,GAR-ARG,GAR-Hierarchical1}. The last one focuses on a crowd of people while the former two commonly pay attention to the videos containing only one or a few people. 

In this paper, we focus on a more comprehensive human activity understanding in the crowded scenes
% More motivation here
-- we are interested in not only the overall activity of the crowd in the scene, but also the instance action of each person (referred to as subject in this paper) and the social activity among a subset of subjects in the crowd.
To achieve such a \textit{comprehensive and multi-granular  human activity understanding}, we propose a new problem -- \textbf{P}anoramic Human \textbf{A}ctivity \textbf{R}ecognition (PAR) that integrates these three sub-tasks. As shown in Fig.~\ref{fig:example} (e), instance action recognition (Task I) aims to recognize the action of each subject in the scene, social group activity recognition (Task II) aims to divide the crowd into social groups and recognize the activity of each group, and global activity recognition (Task III) depicts the general and abstract activity according to the majority of the people. {The proposed PAR problem studies the human activity understanding in multiple spatial granularity (individual, group and global) with different levels of activity categories.}
%We believe that the proposed PAR can better analyze the human activity for a crowd of people, e.g., that in the multi-person social scene and dense crowd monitoring scene.
% task: 
% scene: social life
% motivation: integration complementary

The three tasks in PAR are complementary to each other and the solution of one task may  benefit the others. For example, Task I recognizes the atomic action of each individual, which provides useful information for Task II of social interaction recognition, e.g., a person is talking and a person is listening may indicate the social activity of conversation if they are facing each other. Similarly, the overall activity recognition can be better achieved if we know all the individual actions and social group activities in the scene. On the contrary, the overall activity recognition provides useful priors to  assist the social group activity and individual action understanding. In this paper, we aim to develop a joint framework to simultaneously address these three tasks. Compared to previous human activity understanding tasks, the proposed PAR is more challenging. 
{
A key problem is to establish a unified framework that can jointly handle all the sub tasks together, rather than address them severally or one after another. 
For this purpose, we expect to excavate and leverage the dependence among the different tasks and make them promote each other.}
%The whole network is implemented in an one-stage end-to-end manner that simultaneously achieve the three sub-tasks.

In this paper, we develop a one-stage end-to-end  hierarchical graph network  with a nested graph structure for PAR, {in which the nodes at different hierarchies fitly represent the individual, group and global activities at different levels.}
Specifically, we first build a  graph network by modeling each individual subject as a graph node. We then propose an AiO (all in one) feature aggregation module to aggregate individual feature nodes to group nodes in a bottom-to-up way. Similarly, the individual and group nodes are further aggregated into the global node. We further use a top-to-down feedback strategy for representation fusion and enhancement.
With the hierarchical network architecture, we apply the multi-level supervisions for the multiple tasks in our problem.

The main contributions of this work are summarized as below:

1. We propose a new problem of {P}anoramic Human {A}ctivity Recognition (PAR), which aims to simultaneously recognize the individual human actions, social group activities and global activity in a crowded multi-person scene.

2. We develop a one-stage framework with hierarchical graph network that can effectively and  collectively represent and model the activities  at different level of granularities and mutual relations for multiple people in the scene. 

3. We build a new video benchmark by adding new activity annotations and evaluation metrics to an existing dataset, for the proposed PAR. Experimental results verify the rationality of the proposed problem and the effectiveness of our method. We release the benchmark and code to the public at~\href {https://github.com/RuizeHan/PAR}{\textcolor{magenta}{https://github.com/RuizeHan/PAR}}.

%-------------------------------------------------------------------------
% Background
\section{Related Work}

\textbf{Human action recognition and localization.}
Human action understanding is a fundamental task in computer vision. 
Early works mainly focus on the task of human action recognition, which takes a video including a human with specific action as input and aims to recognize the action category. This task can be also regarded as a video classification problem.
Existing methods for action recognition can be divided into two categories, i.e., the appearance based~\cite{2016Temporal,Zhou_2018_CVPR,Du_2018_ECCV,Diba_2018_ECCV} and the skeleton-based~\cite{vemulapalli2014human,Huang_2017_CVPR,Si_2018_ECCV,Friji_2021_ICCV} methods.
Recent studies started to focus on the action localization (also called action segmentation) task, including temporal action localization and spatial-temporal action localization.
The former is defined to localize the temporal duration of the action in untrimmed videos and then recognize the action category~\cite{Sf-net,zhang2020multi}. The latter not only recognizes the action and localize its duration but also provides the spatial location (in terms of a bounding box) of the corresponding actor~\cite{slowfast,tang2020asynchronous,wu2020context,pan2021actor,dataset-multisports}.
%This task is obviously more challenging and practical, which has drawn more attention recently. 
However, the video data used in these tasks are usually collected from the actions performed by the actors in the laboratory or from the movies/website videos, e.g., UCF~\cite{2012UCF101}, DALY~\cite{dataset-DALY}, Hollywood2tubes~\cite{dataset-Hollywood2tubes}, and AVA~\cite{dataset-ava}. 
Among them, the videos commonly contains only one or very few humans, and the actors usually occupy the main part of the picture in each frame.
{Differently, this paper is focused on the activity understanding in the crowded scenes, which is more practical in many applications, e.g., video surveillance and social analysis, in the real world.}

\textbf{Human interaction recognition.}
Compared to the human action recognition, the study of human-human interaction recognition is less studied.
Existing human interaction recognition mainly focuses on the interactive activity involving two subjects, e.g., shaking, hugging, which occurs more in the scenarios analysis and video surveillance~\cite{Ryoo2009Spatio,Gemeren2016Spatio,Yun2012Two}.
Similar to human action  recognition, most works  on interaction recognition  focus on the videos collected from movies~\cite{Marszalek2009Actions} and TV shows~\cite{Patron2012Structured}.
Some recent works begin to study the human interaction in the multi-person scenes.
For example, in~\cite{zhao2020human}, a new problem of spatial-temporal human-human interaction detection  is studied  in a crowded scene.
More comprehensive introduction to human interaction recognition can be found in a recent survey paper~\cite{2019Analyzing}.
{The classical human interaction as discussed above commonly considers the interactive activities involving two humans. 
Differently, in this paper we detect the social activities where the exact number of involved persons is priorly unknown -- we first divide the people in the scene into human (social) groups with different sizes and then study the interactions in each group. }

{\textbf{Social group/activity detection.}
	Social group detection task aims to divide a crowd of people into different (sub-)groups by the social activities or relations.
	Early methods for this task include the group-based methods without considering each individual person~\cite{shao2014scene,feldmann2010tracking}, the individual-based methods aggregating the information of all individual subjects~\cite{chang2011probabilistic,solera2015socially,ge2012vision} and the combined methods considering both of them~\cite{pang2011detection,bazzani2012decentralized}.
	Recently, several deep learning based methods~\cite{fernando2018gd,GAR-joint-learning,wang2020panda} are developed for the group detection task.
	In the recent PANDA benchmark~\cite{wang2020panda}, human social interaction is treated as auxiliary task for group detection in the crowded scenes.
	Also, a couple of recent works~\cite{GAR-joint-learning,Ehsanpour2021JRDBActAL} aims to detect the social sub-groups in the multi-person scenes and meanwhile recognize the social activity in each sub-group.
	In~\cite{GAR-joint-learning,Ehsanpour2021JRDBActAL}, the social activity in each group is simply regarded as the individual action that is performed by most humans in this group -- this is not practical in many real-world cases.
	In this paper, the task II is focused on the social group detection and its activity recognition, which is complementary to the other two tasks for more comprehensive multi-human activity understanding.
}

\textbf{Group activity recognition.} 
Group activity recognition (GAR) is another task for human activity understanding, which aims to recognize the activity for a group of people.  
Early researches directly take the video recording the activity of a group of people as input, and output the activity category~\cite{Shu_2017_CVPR,Ibrahim_2018_ECCV}, which is more like a video classification task. 
Recent works found that the individual human actions and the human-human interactions can help GAR. 
This way, several methods begin to include the individual action labels as auxiliary supervision for the GAR task~\cite{GAR-Hierarchical1,GAR-joint-learning}.
Several other methods~\cite{Ibrahim_2018_ECCV,GAR-ARG} propose to model the relations among multiple actors for better representation in the GAR task.
{This problem is actually the task III, i.e., the global activity recognition, in the proposed PAR. Difference lies in that we also simultaneously handle the other two tasks of the individual and social-group activity recognition. }

Overall, the proposed PAR problem studies the human activity understanding in {{multiple spatial granules}} (individual, group and global) with {{different levels} of activity categories}, which is underexplored and has important applications.

\section{The Proposed Method}

\subsection{Overview}
	In this work, we aim to jointly learn the individual actions, social group activities (based on the group detection) and the global activity in a unified one-stage framework.
	This way, we propose a hierarchical graph network that can well represent and model the multiple tasks. The architecture of the whole network is shown in Fig.~\ref{fig:framework}. 

	Specifically, given a video recording a crowded scene, we extract the feature of each subject using a deep neural network.
	{For each frame, we first model all the subjects appearing in the scene as a graph, in which each individual subject (feature) is modeled as a graph node, and each edge encodes relation between two subjects. 
	Based on this graph, we propose a novel hierarchical graph network architecture. The proposed method can aggregate the \textit{individual nodes} into the \textit{group nodes} using the proposed bottom-to-up AiO (all in one) feature aggregation module. Similarly, the individual and group nodes are further aggregated into a \textit{global node}. We also apply a top-to-down feedback strategy to further boost the mutual promotion among different tasks through the hierarchical network.
	We will elaborate on the proposed method in the following subsections.
	
%	\vspace{-20pt}
	\begin{figure*}[ht!] 
		\centering 
		\includegraphics[width=0.925\linewidth]{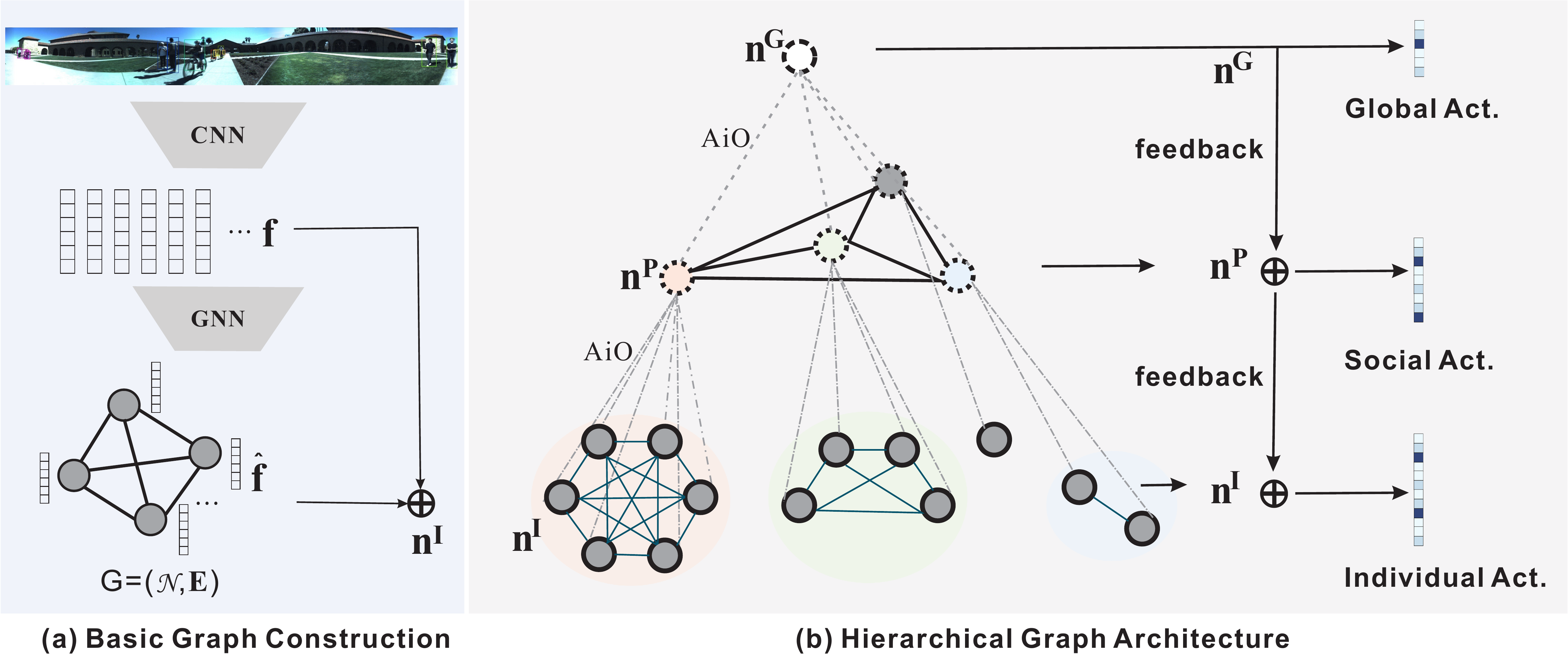}
		\caption{Illustration of the framework of our method, which is mainly composed of the basic graph construction (as discussed in Section~\ref{sec:graph}), and the hierarchical graph network (as discussed in Section~\ref{sec:hier}).}
		\label{fig:framework}  
	\end{figure*}
%\vspace{-20pt}
		
	\subsection{Basic graph construction}
	\label{sec:graph}
	We first build a fully-connected graph $G = (\mathcal{N}, \mathbf{E})$ to represent the $N$ subjects in the scene, in which $\mathcal{N}$ denotes the node set and $\mathbf{E}$ encodes the relations for each pair of nodes.
	
	\textbf{Feature extraction.} 
	%{\ding{182} Appearance feature.}
	Similar to many previous work for action recognition~\cite{GAR-ARG}, we employ the classical CNN network, i.e., Inception-v3~\cite{Inceptionv3} followed by the RoIAlign technique~\cite{RoIAlign}, to extract the deep appearance features of each subject and resize them into the same size.
	%After that, we apply multiple 3D convolutional layers with a kernel size of $3 \times 1 \times 1$ to aggregate the temporal information over the $N$ frames in a segment. 
	After that, we apply a fully connected (FC) layer to obtain the 1,024-dimension appearance feature for each subject.
	In the following, we denote the appearance feature of the subject $u$ as $\mathbf{f}_u \in \mathbb{R}^{1,024}$.
	We use the GCN (Graph Convolutional Network) for the basic graph construction, which contains two parts, i.e., graph edge representation and graph node representation.
	
	\textbf{Graph edge representation.}
	We use affinity matrix $\mathbf{E} \in \mathbb{R}^{N \times N}$ to learn the affinities among all the nodes in the graph, which is updated according to the node features as
	\begin{equation}
	\label{eq:updateA}
	e_{u,v} = \mathrm{F_1}({\mathbf{f}}_{u}) \cdot (\mathrm{F_2}({\mathbf{f}}_{v}))^{\mathrm{T}},  u, v \in \mathcal{N},
	\end{equation}
	where $\textbf{f}_u$ and $\textbf{f}_v$ denote the feature vectors of nodes $u$ and $v$, respectively. 
	$\mathrm{F_1}$ and $\mathrm{F_2}$ denote the MLP (multilayer perception) networks with the same structure but unshared parameters. The output vectors are multiplied into an affinity weight $e_{u,v}$, and the affinity matrix ${\textbf{E}} = [e_{u,v}]_{u,v}\in \mathbb{R}^{N \times N}$ encodes the affinity weights among all node pairs. We finally apply a softmax operation on each row of  ${\textbf{E}}$ for normalization, since it is used as the weights for graph node representation updating.
	
	\textbf{Graph node representation.} 
	We then update graph node feature ${\textbf{f}}_{u}$ through all the connected nodes  weighted by the affinity matrix as
	
	\begin{equation}
	\label{eq:updatex}
	{\hat{\textbf{f}}}_{u} =   \mathrm{F_n}( \sum\nolimits_{v}e_{u,v} {\textbf{f}}_{v}) , u, v \in \mathcal{N},
	\end{equation}
	where $\mathrm{F_n}$ denotes the node update network.
	Finally, we take the updated feature $\hat{\textbf{f}}_{u} $ together with the residual connection to the original feature ${\textbf{f}}_{u}$ as the \textit{individual node representation} $\mathbf{n}^\mathrm{I}_u$, i.e., 
	\begin{equation}
	\label{eq:node}
	\mathbf{n}^\mathrm{I}_u = {\textbf{f}}_{u} \oplus \hat{\textbf{f}}_{u} ,
	\end{equation}
	which embeds both the original individual feature from ${\textbf{f}}_{u}$ and the surrounding-aware information from $\hat{\textbf{f}}_{u} $, as shown in Fig.~\ref{fig:framework}(a).
	This way, we have finished the basic graph construction.
	
	\subsection{Hierarchical Graph Network Architecture}
		\label{sec:hier}
	\subsubsection{Bottom-to-up aggregation (B2U).} After constructing the graph with each subject as a node, we propose to build a hierarchical graph network to model the different-granularity activities in the proposed problem.
	
	\textit{Individual to group aggregation.}  We first consider the {human social group.} We aim to establish the human relation matrix $\mathbf{R} \in \mathbb{R}^{N \times N}$ to represent the human social relation among the subjects, which is calculated from two aspects, i.e., the affinity matrix $\mathbf{E}$ in the basic graph and the distance-aware affinity matrix.
	Specifically, we first calculate the spatial distance matrix $\mathbf{D} \in \mathbb{R}^{N \times N}$ to encode the spatial distance between each two subject as 
	\begin{equation}
	\label{eq:dis}
	{\mathbf{D}}(u,v) = \frac{\sqrt{(x_u-x_v)^2+(y_u-y_v)^2}}{\sqrt{S_u+S_v}},
	\end{equation}
	where $x_u$, $y_u$ denote the coordinate of subject $u$ (midpoint at bottom edge of the bounding box), $S_u$ denotes the area of the bounding box of subject $u$. Here we take the bounding box area into consideration since the principle of near-large and far-small during imaging.
	We then get the relation matrix $\mathbf{R}$ as
	\begin{equation}
	\label{eq:R}
	\mathbf{R} = \lambda \mathbf{E} \odot \bar{\mathbf{D}} \oplus (1-\lambda) \breve{\mathbf{D}},
	\end{equation}
	where $\breve{\mathbf{D}} = \textrm{sigmod}(\frac{1}{\mathbf{D}})$ denotes the distance-aware affinity matrix. Similar with~\cite{GAR-ARG}, we additionally apply a distance mask $\bar{\mathbf{D}}$ on $\mathbf{E}$ to filter the connections between two subjects that are far from each other with a threshold $\rho$. The distance mask matrix $\bar{\mathbf{D}}$ is computed as
	\begin{equation}
	\label{eq:Dmask}
	\bar{\mathbf{D}}(x) =
	\left\{ 
	\begin{aligned} 
	&\mathbf{D}(x),  \qquad \textrm{if} \ \mathbf{D}(x) \leq \rho, \\
	&- \infty, \qquad \textrm{if} \ \mathbf{D}(x) > \rho.
	\end{aligned} \right.
	\end{equation}
	
	With the human relation matrix, we can get the human group division results through a post-processing method, e.g., a clustering algorithm. 
	As shown in Fig.~\ref{fig:framework}(b), we then aggregate the individual nodes $\mathbf{n}^\mathrm{I}_u$ in each group $\mathcal{G}_k$ into the group node representation
	\begin{equation}
	\label{eq:groupnode}
	\mathbf{n}^\mathrm{P}_k = \mathrm{AiO}(\mathbf{n}^\mathrm{I}_u \ | \  u \in \mathcal{G}_k),
	\end{equation}
	where $\mathrm{AiO}(\cdot)$ denotes the proposed all in one (AiO) aggregation module, which will be discussed in detail later. 
	
	\textit{Group to global aggregation.} Similar to the individual to group aggregation, as shown  in Fig.~\ref{fig:framework}(b), we also aggregate the individual and group nodes into a global node, as
	\begin{equation}
	\label{eq:p2g}
	\mathbf{n}^\mathrm{G} = \mathrm{AiO}(\mathbf{n}^\mathrm{I}_u, \mathbf{n}^\mathrm{P}_k \ | \ u \notin  \forall \mathcal{G}, \forall k),
	\end{equation}
	where we take both the individual node $\mathbf{n}^\mathrm{I}_u$ not in any group $\mathcal{G}$, and the aggregated group nodes $\mathbf{n}^\mathrm{P}_k$ in Eq.~\eqref{eq:groupnode}, as the units for aggregation.
	This global node can be used for representing the global activity of all subjects in the scene.
	
	\textit{All in one (AiO) aggregation module.}
	We then present the all in one (AiO) aggregation module used in the above.
	Take the individual to group aggregation using the AiO module in Eq.~\eqref{eq:groupnode} for example.
	We assume the individual nodes $ \mathbf{n}^\mathrm{I}_u, u \in \mathcal{G}_k$ in a group $\mathcal{G}_k$ are aggregated as a group node, we first build a local GCN to get the aggregation matrix $\mathbf{W} \in \mathbb{R}^{|\mathcal{G}_k| \times |\mathcal{G}_k|}$, which is similar with the graph affinity matrix in Eq.~\eqref{eq:updatex}, and $|\mathcal{G}_k|$ denotes the number of subjects in this group.
	We then accumulate the values in each column of $\mathbf{W}$ and get a weight vector $\mathbf{w} \in \mathbb{R}^{1 \times |\mathcal{G}_k|} $.
	We finally aggregate the individual node features with this weight vector as
	\begin{equation}
	\mathbf{n}^\mathrm{P}_k = \sum_u ({w}_u \mathbf{n}^\mathrm{I}_u \ | \  u \in \mathcal{G}_k),
	\end{equation}
	where $\mathbf{n}^\mathrm{P}_k$ denotes the \textit{aggregated node feature} of group $\mathcal{G}_k$, ${w}_u$ is the $u$-th element in $\mathbf{w}$.
	
	The group to global aggregation in Eq.~\eqref{eq:p2g} takes both the individual and group nodes as input and outputs a global node, which can be similarly achieved by the above AiO module.
	
	\subsubsection{Top-to-down feedback (T2D).}
	We use the bottom-to-up hierarchical graph network to model the multi-granular activities. To further improve the mutual promotion among different tasks, we apply a top-to-down feedback strategy.
	Specifically, as shown in Fig.~\ref{fig:framework}(b), we integrate the group node feature together with the individual/group node feature for the final representation and action category prediction as
	\begin{equation}
	\label{eq:T2D}
	\begin{aligned}
	\mathbf{a}^\mathrm{I}_u  = \mathrm{F_i} (\mathbf{n}^\mathrm{I}_u,\mathbf{n}^\mathrm{G}), \quad
	\mathbf{a}^\mathrm{p}_k  = \mathrm{F_p} (\mathbf{n}^\mathrm{P}_k,\mathbf{n}^\mathrm{G}), \quad
	\mathbf{a}^\mathrm{G}  = \mathrm{F_g} (\mathbf{n}^\mathrm{G}),
	\end{aligned}
	\end{equation}
	where $\mathbf{a}^\mathrm{I}_u$, $\mathbf{a}^\mathrm{p}_k$ and $\mathbf{a}^\mathrm{G}$ are the predicted individual, social (group) and global activities, respectively, $\mathrm{F_i}$, $\mathrm{F_p}$, and $\mathrm{F_g}$ are the corresponding readout functions implemented by the MLP neural networks.
	
	\subsubsection{Multi-level multi-task supervisions.}
	We use the multi-level losses as supervisions for the multiple tasks in our problem. The total loss is defined as  
	\begin{equation}
	\label{eq:loss}
	\begin{aligned}
	\mathcal{L} & = \mathcal{L}_i + \mathcal{L}_p + \mathcal{L}_g + \mathcal{L}_d \\
	& = \textstyle \sum_u \mathrm{L}(\mathbf{a}^\mathrm{I}_u, \tilde{\mathbf{a}}^\mathrm{I}_u) + \sum_k \mathrm{L}(\mathbf{a}^\mathrm{P}_k,\tilde{\mathbf{a}}^\mathrm{P}_k) + \mathrm{L}(\mathbf{a}^\mathrm{G},\tilde{\mathbf{a}}^\mathrm{G}) + \mathrm{L}(\mathbf{R},\tilde{\mathbf{R}}),
	\end{aligned}
	\end{equation}
	where $\mathcal{L}_i$, $\mathcal{L}_p $, $\mathcal{L}_g$, $\mathcal{L}_d$ denote the losses for the individual, social group, global activity recognition, and the group detection tasks, respectively. The notations with $\tilde{\cdot}$ denote the corresponding ground-truth labels.a $\tilde{\mathbf{R}} \in \mathbb{R}^{N \times N}$ is the human group relation matrix taking the values of $0$ or $1$, where 1 denotes the corresponding two subjects are in the same group.
	
	\subsection{Implementation Details}
	
	\textbf{Network details.}
%	We provide some implementation details in the proposed method. 
	The MLP networks $\mathrm{F_1}$ and $\mathrm{F_2}$ in Eq.~\eqref{eq:updateA}, and $\mathrm{F_n}$ in Eq.~\eqref{eq:updatex} are all implemented by single-layer FC networks. 
	The parameter $\lambda$ in Eq.~\eqref{eq:R} is set as 0.5.
	The parameter $\rho$ in Eq.~\eqref{eq:Dmask} is set as the width of the input image with a ratio of 0.2.
	The readout function $\mathrm{F_i}$, $\mathrm{F_p}$  in Eq.~\eqref{eq:T2D}  are implemented by the three-layer FC networks, and $\mathrm{F_g}$ by a two-layer FC network.
	We use the binary cross entropy as the loss function in Eq.~\eqref{eq:loss}.
	As a new problem, in this work, we use the annotated human bounding boxes as input to alleviate the interference from the false human detection, which is common for the crowded scene in our problem.
	In our method, we do not integrate the temporal information of the individual and group along the video, since this need the multi-object tracking and group evolution detection results as auxiliaries. The challenging scenes make them not easy to be obtained, and the involved errors will have an impact on the main task.
	
	\textbf{Network training.}
	During the training stage, we use the ground-truth human group division for individual to group aggregation in Eq.~\eqref{eq:groupnode}.
	We implement the proposed network with the PyTorch framework on the GTX 3090 GPU.
	The batch size is set as 4 in the experiments.
	We use stochastic gradient descent (SGD) algorithm with Adam method for training the network, which is trained for about 50 epochs with the learning rate of $2 \times 10^{-5}$ to be convergent.
	
	\textbf{Inference.} During the inference stage, we get the human group division results through a post-processing method.
	Specifically, with the predicted human relation matrix $\mathbf{R}$, we apply a self-tuning spectral clustering algorithm~\cite{2004Self} that uses $\mathbf{R}$ as input and automatically estimates the number of clustering groups, to obtain the  group detection results.
	
	\section{Experiments}
	
	\subsection{Datasets}
	
	Previous datasets for human action or group activity recognition can not meet the requirements of the proposed problem. We build a new benchmark for the proposed task. 	
	The proposed dataset is based on a state-of-the-art dataset JRDB~\cite{martin2021jrdb} for 2D/3D person detection and tracking, which uses a mobile robot to capture the 360$^{\circ}$ RGB videos for crowded multi-person scenes%	\footnote{The input videos for the proposed panoramic human activity recognition task include but not limited to the 360$^{\circ}$ panoramic video.}
	, such as those at the campus, canteen, and classroom, etc.
	JRDB has provided the human bounding boxes with IDs as annotations. 
	Based on it, a more recent dataset JRDB-Act~\cite{Ehsanpour2021JRDBActAL} adds the annotations of individual human action and social group detection. The group activity label in JRDB-Act is simply defined as the combination of the involved individual action labels or the selection of them considering the occurrence frequency in the group, which seems not reasonable enough in practice. In PAR problem, the group activity is related to the individual actions but their label candidate sets are different. For example, for a two-person conversation scene, in~\cite{Ehsanpour2021JRDBActAL} the group activity is labeled as `listening to someone, talking to someone’, which is the combination of the contained individual action labels.
	In our setting, we label such group activity as ‘chatting’, which is regarding to the whole group but not each individual.
	This way, we inherit the human detection annotations in JRDB and the individual action and group division annotations in JRDB-Act, and further manually annotate the social group activities and global activities for the proposed Panoramic Human Activity Recognition (PAR) task, which constitutes a new dataset -- JRDB-PAR.

	\begin{figure*}[ht!] %\vspace{-10pt}
		\centering 
		\includegraphics[width=1\linewidth]{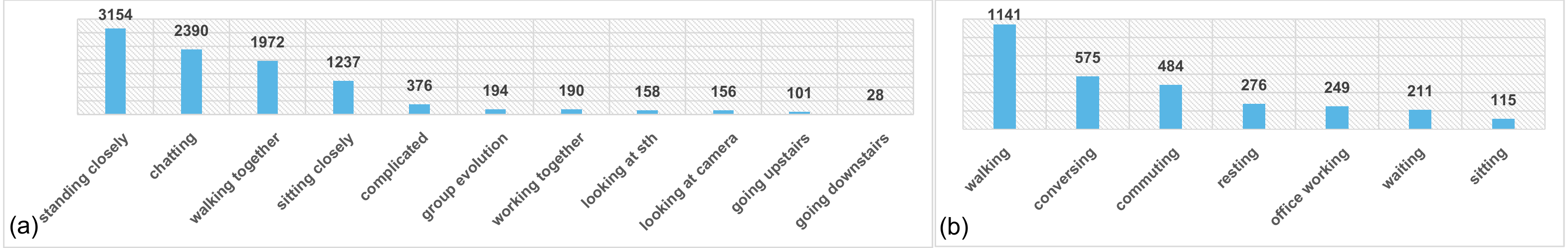}
		\caption{Data distributions of social (a) and global activities (b) in JRDB-PAR. }
		\label{fig:data}  
	\end{figure*} 
	
	JRDB-PAR contains 27 categories of individual actions, e.g., walking, talking, which is same as JRDB-Act, and 11 categories of social group activities and 7 categories of global activities. The distributions of social/global activities are shown in Fig.~\ref{fig:data}.
	In total, JRDB-PAR includes 27 videos, which are splitted as 20 for training and 7 for testing, following the training/validation splitting in JRDB dataset.
	According to statistics, JRDB-PAR contains 27,920 frames with over $628 k$ human bounding boxes.
	Following the setting in~\cite{Ehsanpour2021JRDBActAL}, we select the uniformly sampled key frames (one key frame in each 15 frames) for annotation and evaluation, which is same as previous classical activity recognition datasets like CAD~\cite{dataset-CAD}, volleyball~\cite{dataset-volleyball}.
	The numbers of frames, groups and subjects in the whole dataset and key frames are summarized in Table~\ref{tab:data}.
	Other details for annotated labels, e.g., the number of labels for individual actions, social activities and global activities are also shown in Table~\ref{tab:data}. Note that, we adopt the multi-class labels for the activity annotation, i.e., each individual/group/frame is with one or multiple activity labels.
		
	%number of key frames/individual actions/groups/group activities/global activities
	%1) Video and annotation.  Video resource, action category, group annotation
	%
	%2) Dataset statistics. Data scale, Split
	%
	%CAD-Social
%\vspace{-10pt}
	\begin{table*}[htbp]
	\caption{{Statistics results of the panoramic human activity recognition dataset.}}
	\label{tab:data}  %\vspace{-10pt}
	\begin{spacing}{1.05}
		\centering
		\renewcommand\tabcolsep{5pt}
		\footnotesize
		\begin{tabular}[*c]{l|ccc|ccc|ccc}
			\Xhline{1pt}
			%\rowcolor{mygray}  
			\multirow{2}{20pt}{} &\multicolumn{3}{c}{{All Frames}} &\multicolumn{3}{c}{{Key Frames}}    &\multicolumn{3}{c}{{Activity Labels}} \\   \cline{2-10} 
			%\rowcolor{mygray}
			&Frm. &Gro.  &Sub.
			&Frm.  &Gro.  &Sub. & Indiv.  & Social  & Global \\ \hline  \hline
			Train.  &21,724	&86,949		&467$k$ &1,439		&5,163	&27$k$ &57,341	&7,874	&2,316 \\
			\rowcolor{mygray}		
			Test. &6,196	&31,035		&160$k$  &411		&1,709 &9$k$	 &18,019	&2,082	&735\\		
		    All 	&27,920		&117,984 &628$k$	&1,850  &6,872 	&36$k$	 &75,360	&9,956	&3,051\\	   
			\Xhline{1pt}      
		\end{tabular} 
	\end{spacing}  
\end{table*} %\vspace{-20pt}

	\subsection{Metrics}
	
	\textbf{Protocol I.} To evaluate the \textit{individual action detection}, following the previous work~\cite{godbole2004discriminative} for multi-label classification task, we adopt the commonly used metrics – precision, recall and F$_1$ score (denoted as $\mathcal{P}_i$, $\mathcal{R}_i$, and $\mathcal{F}_i$) as the evaluation metrics, which measure the individual action classification accuracy for each instance in the testing dataset.
	
	\textbf{Protocol II.} \textit{Social (group) activity recognition} includes the group detection and the activity category recognition.
	For the group detection, we use the classical Half metrics for performance evaluation, where the group member IoU $>$ 0.5 in the predicted group and ground-truth group is taken as the true detected group, which is followed by the general protocol in group detection task~\cite{wang2020panda}. Note that, we only consider the groups containing more than one subjects in this protocol.
	For the true detected group under the above metric, we further consider their group activity recognition results. The true detected groups with the correct activity category prediction are taken as the true social group activity predictions. Under this protocol, we calculate the precision, recall and F$_1$ score (denoted as  $\mathcal{P}_p$, $\mathcal{R}_p$, and $\mathcal{F}_p$) as the social activity recognition metrics.
	
	\textbf{Protocol III.} \textit{Global activity recognition} can be also taken as a multi-label classification problem, and we apply the precision, recall, and F$_1$ score (denoted as  $\mathcal{P}_g$, $\mathcal{R}_g$,  $\mathcal{F}_g$) for evaluation.
	
	The overall metric for the panoramic activity detection task is the comprehensive results of the above three metrics. Here we simply compute the average value of the above F$_1$ scores, i.e., $\mathcal{F}_a   = \frac{1}{3} (\mathcal{F}_i  + \mathcal{F}_p  + \mathcal{F}_g)$ as the overall F$_1$ score.
	
	%\begin{equation}
	%\label{eq:metric}
	%\begin{aligned}
	%\mathcal{F}_a  & = \frac{1}{3} (\mathcal{F}_i  + \mathcal{F}_p  + \mathcal{F}_g) \\ 
	%%\mathcal{F}  & = \frac{1}{N} (N_i \cdot \mathcal{F}_i  + N_p \cdot \mathcal{F}_p  + N_g \cdot \mathcal{F}_g) \\
	%%\mathcal{F}  & = \frac{1}{2N} (N - N_i) \cdot \mathcal{F}_i  + (N-N_p) \cdot \mathcal{F}_p  + (N-N_g) \cdot \mathcal{F}_g \\
	%\end{aligned} 
	%\end{equation}
	
	\subsection{Results}
	
	\textbf{Comparison methods.} For the proposed PAR problem, we can not find many methods for direct comparison.
	We try to include more state-of-the-art methods with necessary modifications as the comparison methods.
	ARG is a well-know method~\cite{GAR-ARG} for group activity recognition, which builds a learnable graph structure to model the human relation graph. SA-GAT~\cite{GAR-joint-learning} proposes to learn the human actions, sub-group and group activities together as a multi-stage multi-task problem, 
	which uses the classical group activity recognition dataset CAD~\cite{dataset-CAD} with new sub-group activity annotations for evaluation. 
	JRDB-Base~\cite{Ehsanpour2021JRDBActAL} is the baseline method for the JRDB-Act dataset, which is similar to~\cite{GAR-joint-learning} and uses a progressive multi-loss strategy to recognize the different types of activities. 
	Note that, the original version of ARG can not provide the group detection and social group activity recognition.
	We apply the group clustering algorithm used in our method and previous works~\cite{GAR-joint-learning}, on the relation matrix generated by ARG for group detection.
	With the group detection results, we further employ the feature fusion method in~\cite{Ehsanpour2021JRDBActAL} on ARG for the social group activity recognition task.
	
	We also include several state-of-the-art methods for individual and group activity recognition for comparison, i.e., AT~\cite{AT}, SACRF~\cite{Pramono2020EmpoweringRN}, Dynamic~\cite{yuan2021visualcontext}, HiGCIN~\cite{yan2020higcin}. These methods can not provide the group detection and social activity recognition results, which also do not include the relation matrix in ARG.
	For comparison, we additionally provide the ground-truth group detection results as input to these methods and employ the feature fusion strategy in~\cite{Ehsanpour2021JRDBActAL} to obtain the social group activity recognition results.
%	\vspace{-10pt}
	\begin{table*}[htbp]
		\caption{{Comparative results of the panoramic human activity recognition (\%).}}
		\label{tab:res}  %\vspace{-10pt}
		\begin{spacing}{1.05}
			\centering
			\renewcommand\tabcolsep{5pt}
			\footnotesize
			\begin{tabular}[*c]{l|ccc|ccc|ccc|c}
				\Xhline{1pt}
				%\rowcolor{mygray}  
				\multirow{2}{30pt}{Method} &\multicolumn{3}{c}{{Individual Act.}} &\multicolumn{3}{c}{{Social Act.}}    &\multicolumn{3}{c}{{Global Act.}} &Overall \\   \cline{2-11} 
				%\rowcolor{mygray}
				&$\mathcal{P}_i$  &$\mathcal{R}_i$  &$\mathcal{F}_i$
				&$\mathcal{P}_p$   &$\mathcal{R}_p$  &$\mathcal{F}_p$ 
				&$\mathcal{P}_g$   &$\mathcal{R}_g$  &$\mathcal{F}_g$ 
				& $\mathcal{F}_a$ \\ \hline  \hline
				ARG~\cite{GAR-ARG} &39.9 	&30.7 &33.2 &8.7 &8.0 &8.2 &63.6 &44.3 &50.7 &30.7\\
				\rowcolor{mygray}		
				SA-GAT~\cite{GAR-joint-learning} &44.8 &40.4 &40.3 &8.8 	&8.9 &8.8 	&36.7 &29.9 &31.4 &26.8\\		
				JRDB-Base~\cite{Ehsanpour2021JRDBActAL} &19.1	&34.4	&23.6	&14.3	&12.2	&12.8	&44.6	&46.8	&45.1	&27.2\\	
				\rowcolor{mygray}			
				Ours   &51.0 	&40.5 	&43.4	&24.7	&26.0	&24.8	&52.8 	&31.8 	&38.8	&35.6 \\
				\hline   \hline 			
				AT$^*$\cite{AT} &38.9 &	33.9 &34.6	&32.5 	&32.3 &32.0	&21.2 	&19.1 &19.8	&28.8 \\
				\rowcolor{mygray}	
				SACRF$^*$\cite{Pramono2020EmpoweringRN} &31.3 &23.6   &25.9	&25.7 	&24.5 &24.8	&42.9 	&35.5  &37.6	&29.5  \\
				Dynamic$^*$\cite{yuan2021visualcontext} &40.7 	&33.4 &35.1	&33.5 	&30.1 &30.9	&37.5 &27.1 &30.6	&32.2 \\
				\rowcolor{mygray}	
				HiGCIN$^*$\cite{yan2020higcin} &34.6 	&26.4 &28.6		&34.2 	&31.8 &32.2	&39.3 	&30.1 &33.1	&31.3  \\		
				ARG$^*$\cite{GAR-ARG} &42.7 	&34.7 &36.6			&27.4 	&26.1 &26.2			&26.9 	&21.5 &23.3 &28.8 \\
				\rowcolor{mygray}
				SA-GAT$^*$\cite{GAR-joint-learning} &39.6 &34.5 &35.0 &	32.5 	&32.5 &	30.7 	&28.6 	&24.0 	&25.5 	&30.4  \\	
				JRDB-Base$^*$\cite{Ehsanpour2021JRDBActAL} &21.5	&44.9	&27.7	&54.3	&45.9	&48.5	&38.4	&33.1	&34.8	&37.0 \\
				\rowcolor{mygray}
				Ours$^*$ &54.3 &44.2 &46.9 &50.3 &52.5 &50.1 	&42.1 	&24.5 	&30.3 	&42.4 \\
				\Xhline{1pt}      
			\end{tabular} 
		\end{spacing}  
		{\footnotesize * denotes that we additional provide the ground-truth group detection results as input.}  
	\end{table*} % \vspace{-15pt}
	
	\textbf{Comparative Results.} We show the performance of the proposed method and other comparative methods in Table~\ref{tab:res}. From the top half part, we can first see that the proposed method outperforms the comparative methods on the overall score. More specifically, we can see that the proposed method achieves particularly good performance on the individual and group activity recognition. ARG performs well on the group activity recognition task, which may benefit from its flexible actor relation graphs and sparse temporal sampling strategy. 
	
	We further compare the proposed method with other state-of-the-art methods in the bottom half part of Table~\ref{tab:res}. We can first see that, with the ground-truth group detection as input, the comparison methods perform better for the social activity detection task, which outperform the proposed method. But the overall scores of most comparison methods are still lower than our method. 
	If we also provide the ground-truth group detection in our method, i.e., `Ours$^*$', the performance will be further improved as shown in the last row.

	\subsection{Ablation Study}
	We evaluate the variations of our method by removing some components.\\
	$\bullet$ w/o $\mathbf{f}$ / $\tilde{\mathbf{f}}$ : For the individual node representation, we remove the original individual feature ${\textbf{f}}_{u}$ or the surrounding-aware feature $\hat{\textbf{f}}_{u}$ in Eq.~\eqref{eq:node}, respectively. 
	\\
	$\bullet$ w Euclid. dist.: We replace the calculation method of subject distance in Eq.~\eqref{eq:dis} with the commonly used Euclidean distance.
	\\
	$\bullet$ w/o $\breve{\mathbf{D}}$ / $\mathbf{E}$ : We remove the distance-aware matrix $\breve{\mathbf{D}}$ or the edge matrix $\mathbf{E}$ in the graph network in Eq.~\eqref{eq:R}, respectively.
	\\
	$\bullet$ w/o AiO : We replace the AiO aggregation module in Eqs.~\eqref{eq:groupnode} and~\eqref{eq:p2g} with the  max-pooling operation used in~\cite{Ehsanpour2021JRDBActAL,GAR-joint-learning}.
	\\
	$\bullet$ w/o G.2I. / G.2P.: We remove the global node feature in the individual (G.2I.) or group (G.2P.) node feature representation in Eq.~\eqref{eq:T2D}, respectively.
	
%	\vspace{-10pt}
	\begin{table*}[htbp]
		\caption{{Ablation study of the proposed method with its variations (\%).}}
		\label{tab:abla} % \vspace{-10pt}
		\begin{spacing}{1.05}
			\centering
			\renewcommand\tabcolsep{5pt}
			\footnotesize
			\begin{tabular}[*c]{l|ccc|ccc|ccc|c}
				\Xhline{1pt}
				
				\multirow{2}{30pt}{Method} &\multicolumn{3}{c}{{Individual Act.}} &\multicolumn{3}{c}{{Social Act.}}    &\multicolumn{3}{c}{{Global Act.}} &Overall \\   \cline{2-11} 
				%\rowcolor{mygray}
				&$\mathcal{P}_i$  &$\mathcal{R}_i$  &$\mathcal{F}_i$
				&$\mathcal{P}_p$   &$\mathcal{R}_p$  &$\mathcal{F}_p$ 
				&$\mathcal{P}_g$   &$\mathcal{R}_g$  &$\mathcal{F}_g$ 
				& $\mathcal{F}_a$ \\ \hline  \hline	
				w/o $\mathbf{f}$ &35.6	&27.2	&29.6	&17.4	&15.9	&16.4	&40.2	&23.8	&29.3	&25.1\\
				\rowcolor{mygray}
				w/o $\tilde{\mathbf{f}}$ &29.3	&21.7	&23.9	&9.8	&9.2	&9.4	&43.1	&27.0	&32.4	&21.9 \\
				\hline
				%			w/o $\bar{\mathbf{D}}$ \\
				w Euclid. dist. &43.2	&32.5	&35.6	&12.7	&12.1	&12.3	&41.6	&24.5	&30.2	&26.0\\
				\rowcolor{mygray} 
				w/o $\breve{\mathbf{D}}$ &34.6	&25.8	&28.2	&11.3	&10.8	&11.0	&24.8	&16.1	&19.0	&19.4\\	
				w/o $\mathbf{E}$ &47.8	&37.8	&40.6	&21.0	&20.9	&20.6	&33.6	&19.5	&24.2	&28.4	\\
				\rowcolor{mygray} 
				w/o AiO &19.8	&14.6	&16.1	&18.8	&16.7	&17.3	&34.2	&30.4	&31.3	&21.6 \\
				\hline			
				w/o G.2I. &31.1	&22.9	&25.3	&12.5	&10.9	&11.4	&75.4	&45.0	&55.2	&30.6\\
				\rowcolor{mygray}
				w/o G.2P. &36.4	&27.5	&30.0	&11.9	&10.4	&10.9	&78.4	&46.8	&57.3	&32.7\\
				%				w/o G.2I.P. &48.2	&37.3	&40.4	&21.3	&21.9	&21.2	&44.3	&26.4	&32.4	&31.3\\	
				\hline \hline
				Ours   &51.0 	&40.5 	&43.4	&24.7	&26.0	&24.8	&52.8 	&31.8 	&38.8	&35.6 \\	
				%ours30 & 55.7 & 47.0 & 51.0 & 58.8 & 52.6 & 44.3 & 48.1 & 63.9 & 76.3 & 87.5\\
				\Xhline{1pt}            
			\end{tabular} % \vspace{-10pt}
		\end{spacing}
	\end{table*} 
	
	We show the ablation study results in Table~\ref{tab:abla}. We can first see that original individual feature ${\textbf{f}}_{u}$ and the surrounding node embedded feature $\hat{\textbf{f}}_{u}$ are both important for individual node representation. Especially the latter $\hat{\textbf{f}}_{u}$ is very useful to the social activity recognition. 
	
	Next, in the bottom-to-up (B2U) aggregation stage, we first investigate the effect of the spatial-aware matrix $\breve{\mathbf{D}}$ and the edge matrix $\mathbf{E}$ in relation matrix learning. 
	We can see that $\breve{\mathbf{D}}$ is vital in relation modeling. It is easy to explain that the social relations among the people in the crowd are highly related to the spatial distance. But the simple Euclidean distance between bounding box center performs not well enough. Also, only using the spatial-aware matrix, i.e., w/o $\mathbf{E}$, can not get the performance as using both of them. We then investigate the effectiveness of the AiO module in the proposed B2U aggregation, we can see that the aggregation method AiO is verified to be effective in our method.
	
	Finally, in the top-to-down (T2D) feedback stage,  we can see that the individual action recognition results get worse when removing the global node feature in individual node representation. Similarly, the social activity recognition performance also gets worse without the global node feature embedding. This demonstrates the effectiveness of the T2D feedback strategy in our method.

	\subsection{Experimental Analysis}
	
	\textbf{Individual/global activity recognition.}
	We evaluate the tasks of individual action and global activity recognition, where we follow the original settings in the comparative GAR methods without modification. We show the results in Table~\ref{tab:subtask}. We can see that, although the margin is not very large, the proposed method performs better than the comparative methods on these two tasks.
	We can also find that the performance on JRDB-PAR is generally low%	, which does not exceed 50\% for F$_1$ score. This demonstrates that the dataset is far from saturated
	, where is great potential for more methods to be developed.	
	
%	\vspace{-10pt}
	\begin{table*}[htbp]
		\caption{{Comparisons of the individual action and global activity recognition (\%).}}
		\label{tab:subtask}  %\vspace{-10pt}
		\begin{spacing}{1.05}
			\centering
			\renewcommand\tabcolsep{6pt}
			\footnotesize
			\begin{tabular}[*c]{l|ccc|ccc}
				\Xhline{1pt}
				%\rowcolor{mygray}  
				\multirow{2}{30pt}{Method} &\multicolumn{3}{c}{{Individual Act.}}    &\multicolumn{3}{c}{{Global Act.}}\\   \cline{2-7} 
				%\rowcolor{mygray}
				&$\mathcal{P}_i$  &$\mathcal{R}_i$  &$\mathcal{F}_i$
				&$\mathcal{P}_g$   &$\mathcal{R}_g$  &$\mathcal{F}_g$ 
				\\ \hline  \hline		
				AT~\cite{AT} &36.8 &	30.1 &31.7 	&17.4 	&15.7 &16.1  \\
				\rowcolor{mygray}  
				SACRF~\cite{Pramono2020EmpoweringRN} &39.2 &29.4 &32.2  &	34.8 &	26.2  &28.4  \\
				Dynamic~\cite{yuan2021visualcontext} &46.6 &	37.7 &39.7 	&31.9 	&23.7 &26.4 \\
				\rowcolor{mygray}
				HiGCIN~\cite{yan2020higcin} &36.9 	&30.1 &31.6 		&46.0 	&34.2 &38.0  \\
				\hline \hline
				Ours &51.0 	&40.5 	&43.4 	&52.8 	&31.8 	&38.8 \\
				\Xhline{1pt}            
			\end{tabular} %\vspace{-0pt}
		\end{spacing}
	\end{table*} 
%	\vspace{-10pt}
	
	\textbf{Group detection.} We further solely evaluate the group detection results in Table~\ref{tab:group}. We first use the classical Half metric used in group detection task~\cite{wang2020panda}, i.e., the predicted group with the group member IoU $>$ 0.5 with the ground-truth group is taken as the true detected group, which is denoted as IOU@0.5.
	We extend this metric by increasing the threshold 0.5 to 1 with a step of 0.1, which means improving the criteria bar. We then plot the accuracy curve using each threshold and the corresponding accuracy.
	This way, we can compute the AUC (Area Under the Curve) score as the metric namely IOU@AUC.
	With the predicted and ground-truth binary human group relation matrices ${\bar{\mathbf{R}}}$ and $\tilde{\mathbf{R}}$, where 1 denotes the corresponding two subjects are in the same group, we compute the matrix IOU score as Mat.IOU = $\frac{\sum \mathrm{AND}(\bar{\textbf{R}},{\tilde{\mathbf{R}}})}{\sum \mathrm{OR}(\bar{\textbf{R}},{\tilde{\mathbf{R}}})}$, where $\mathrm{AND}$ and $\mathrm{OR}$ denotes the functions of element-wise logical and/or operations.
	
	As shown in Table~\ref{tab:group}, we first apply two straightforward approaches for comparison. Here `Dis.Mat + LP' denotes that we use the distance matrix among the subjects followed by a label propagation method~\cite{zhan2018consensus} algorithm as the post-processing method to get the group relation matrix. `GNN + GRU' uses a GNN to model the group relations among the subjects followed by a classical GRU model to integrate the temporal information. We also include other methods, i.e., ARG~\cite{GAR-ARG}, SA-GAT~\cite{GAR-joint-learning}, and JRDG-Base~\cite{Ehsanpour2021JRDBActAL}, that can provide the group detection results for comparison.
	We can see that, the proposed method outperforms the comparison methods with a large margin under all metrics.
	
%	\vspace{-12pt}
	\begin{table}[htbp] 
		\caption{Comparative group detection results (\%).}
		\label{tab:group} %\vspace{-10pt}
		\begin{spacing}{1.05}
			\centering
			\renewcommand\tabcolsep{8pt}
			\footnotesize
			\begin{tabular}[*c]{l|ccc}
				\Xhline{1pt}
				&{IOU@0.5} &{IOU@AUC} &{Mat.IOU} \\  \cline{1-4}
				Dis.Mat + LP~\cite{zhan2018consensus} 	& 33.4  & 14.1  & 12.9  \\	
				\rowcolor{mygray}  				
				GNN + GRU		&34.5 	  & 21.7 &20.1 \\
				\hline 
				ARG~\cite{GAR-ARG}	& 35.2   & 21.6	 &19.3\\
				\rowcolor{mygray}
				SA-GAT~\cite{GAR-joint-learning}  &29.1  &20.4  &16.6 \\				
				JRDG-Base~\cite{Ehsanpour2021JRDBActAL}  &38.4  &26.3  &20.6   \\
				\hline   \hline
				\rowcolor{mygray}
				Ours			&\textbf{53.9}  &\textbf{38.1} & \textbf{30.6} \\
				\Xhline{1pt}            
			\end{tabular}
		\end{spacing} 
	\end{table} %\vspace{-30pt}
	
	\section{Conclusion}
	
	In this paper, we have studied a new problem -- panoramic human activity recognition (PAR), which is more comprehensive, challenging and practical than previous action recognition tasks in crowded scene video analysis. 
	To handle this problem, we propose a novel hierarchical neural network composed of the basic graph construction, bottom-to-up node aggregation, and top-to-down feedback strategy, to model the 
	multi-granular activities.
	Based on an existing dataset, we build a benchmark  for this new problem. Experimental results are very promising and verify the rationality of this new problem and the effectiveness of the proposed baseline method. 
	In the future, we plan to integrate the temporal information, e.g., the multiple object tracking~\cite{han2020complementary,han2021multiple} and group evolution detection, into the proposed method.
	An interesting problem maybe to investigate whether we can simultaneously handle the multiple object tracking~\cite{gan2021self} and PAR and make them help with each other.
	
	\textbf{Acknowledgment}. This work was supported by the National Natural Science Foundation of China under Grants U1803264, 62072334, and the Tianjin Research Innovation Project for Postgraduate Students under Grant 2021YJSB174.

	{
		\bibliographystyle{splncs04}
		\bibliography{IEEEabrv,PanoAct}
	}

\end{document}